\pdfoutput=1

\documentclass[11pt]{article}

\usepackage{EACL2023}

\usepackage{times}
\usepackage{latexsym}

\usepackage[T1]{fontenc}

\usepackage[utf8]{inputenc}

\usepackage{microtype}

\usepackage{inconsolata}

\usepackage{graphicx}
\graphicspath{ {./figures/} }

\usepackage{xcolor}

\newcommand\extrafootertext[1]{
    \bgroup
    \renewcommand\thefootnote{\fnsymbol{footnote}}
    \renewcommand\thempfootnote{\fnsymbol{mpfootnote}}
    \footnotetext[0]{#1}
    \egroup
}

\newcommand{\wb}{NLP Workbench}

\title{\wb: Efficient and Extensible Integration of State-of-the-art Text Mining Tools}

\author{Peiran Yao, Matej Kosmajac, Abeer Waheed, \\ {\bf Kostyantyn Guzhva, Natalie Hervieux, Denilson Barbosa}\\
Department of Computing Science\\
University of Alberta\\
{\tt \{peiran, denilson\}@ualberta.ca}}

\begin{document}
\maketitle
\extrafootertext{Camera-ready version for EACL 2023: System Demonstrations.}
\begin{abstract}
    \wb{} is a web-based platform for text mining that allows non-expert users to obtain semantic understanding of large-scale corpora using state-of-the-art text mining models.
    The platform is built upon latest pre-trained models and open source systems from academia that provide semantic analysis functionalities, including but not limited to entity linking, sentiment analysis, semantic parsing, and relation extraction.
    Its extensible design enables researchers and developers to smoothly replace an existing model or integrate a new one.
    To improve efficiency, we employ a microservice architecture that facilitates allocation of acceleration hardware and parallelization of computation.
    This paper presents the architecture of \wb{} and discusses the challenges we faced in designing it.
    We also discuss diverse use cases of \wb{} and the benefits of using it over other approaches.
    The platform is under active development, with its source code released under the MIT license\footnote{\url{https://github.com/U-Alberta/NLPWorkbench/}}.
    A website\footnote{\url{https://newskg.wdmuofa.ca}} and a short video\footnote{\url{https://vimeo.com/801006908}} demonstrating our platform are also available.
\end{abstract}

\section{Introduction}

Text mining, also known as \textit{text analytics} or \textit{text analysis}, is the process where a user interacts with machine-supported analysis tools that transform natural language text into structured data, to gain insights and new knowledge from the text \cite{feldman2006introduction}.
For more than two decades, text mining systems have been built for applications in various domains, such as business intelligence, analytical sociology, and medical sciences \cite{hearst1999untangling}, demonstrating irreplaceable value.
Analysis tools in text mining usually take the form of machine learning (ML) and natural language processing (NLP) models and span a large spectrum of ML and NLP subfields, such as entity linking, sentiment analysis, relation extraction, and text summarization.

Nearly every subfield of NLP involved in text mining has been rapidly evolving in recent years, with records on benchmarks being continuously broken\footnote{\url{http://nlpprogress.com}}.
A positive practice of releasing the code and models to the public has been adopted by a growing number of researchers\footnote{\url{https://paperswithcode.com}} to address reproducibility and accessibility issues in the field\footnote{\url{https://aclrollingreview.org/responsibleNLPresearch/}}.
Despite efforts to make new models more accessible, non-expert users such as digital humanists and business analysts still face entry barriers when trying to apply the latest models. Some salient issues include: (1) heterogenous software stacks required to run the models; (2) non-standardized, inconsistent input and output formats; (3) the lack of user-friendly interfaces to apply the models and visualize the results; and (4) the constraints on computation and networking resources.
We build \wb{} with the goal of addressing these issues and further bridging the gap between state-of-the-art open NLP research and the use of these models and tools in text mining applications by non-experts.

\wb{} is designed with two fundamental principals in mind: for developers and NLP researchers, fast and easy adaptation of off-the-shelf models and tools; and for non-expert users such as sociologists, a user-friendly interface for both document-level and corpus-level analysis.
Following these principles, \wb{} offers the following key features:

\paragraph{Platform} \wb{} unifies corpus management, text mining tools, and visualization in a single platform.
It provides a growing list of models and tools that are based on state-of-the-art research, currently offering functionalities like named entity recognition, entity linking, relation extraction, semantic parsing, summarization, sentiment analysis, and social network analysis.
\paragraph{Interaction} A web interface is included for user interactions with the ability to visualize the model results at document and corpus levels.
Users could choose to interactively apply a model on a given document and have the results saved for future queries, or to apply models in batches on selected documents in a corpus.
\paragraph{Architecture} For development, \wb{} adopts containerization, allowing new models to be added independent of the software stack of existing models.
For deployment, its microservice architecture allows models to be deployed in a distributed way on machines that meet the computing and networking requirements of individual models, enabling horizontal scaling.
\paragraph{Interface} Tools in \wb{} can be accessed in versatile ways. Besides the web interface, non-expert users could import new documents into the platform via a browser extension. For developers and researchers, \wb{} provides RESTful API and remote procedure call (RPC) interfaces for easy integration with other applications and pipelines.
\section{Related Work}
\citet{hearst1999untangling} and \citet{cunningham2002gate} identified three key aspects of an effective text mining system: management of text document collections (corpora), application of text processing algorithms on the collection, and visualization of results.
LINDI \cite{hearst1999untangling} is an early prototype of such a system used for gene function discovery.
GATE \cite{cunningham2002gate}, a framework that is still currently maintained, provides a unified architecture for all three aspects.
Similarly, \wb{} tries to accommodate all three aspects in a single platform.
Our platform uses containerized microservices instead of Java classes for each processing module, which avoids the restrictions of underlying programming frameworks for implementing NLP algorithms.
Voyant Tools \cite{rockwell2016hermeneutica} is another web-based platform that provides tools for corpus analysis and the function to write user-defined scripts.
Their built-in tools are mostly limited to count-based statistics and visualization, while we are integrating large deep learning models.
UIMA \cite{ferrucci2004uima} attempts to define a standard protocol for managing corpora and NLP algorithms. It is more developer-oriented, unlike our application which provides a complete system that can be used for analysis directly by users.

A plethora of NLP toolkits focusing on building NLP pipelines have been developed in the past few decades, including Stanford CoreNLP \cite{manning2014stanford}, OpenNLP\footnote{\url{https://opennlp.apache.org}}, NLTK \cite{bird2009nltk}, spaCy \cite{honnibal2020spacy} and Transformers \cite{wolf2020transformers}.
These toolkits could be used to address the text processing algorithm aspect of text mining systems, but they do not provide a full solution.
The typical strategy to incorporate new models into these toolkits is to re-implement the model in the framework of the toolkit, while we try to re-use, as much as possible, the code and models released by researchers.

Some tools specialize in only the visualization aspect. To name a few, \citet{blloshmi2021spring} and \citet{cohen2021repgraph} built tools for visualizing the results of semantic parsers -- a function that is also provided by our system.

Several libraries and tools are able to manage corpora or models from multiple sources.
For example, Datasets library \cite{lhoest2021datasets} provides an interface to access common NLP datasets, and DataLab \cite{xiao2022datalab} is a platform to examine and analyze datasets.
Transformers \cite{wolf2020transformers} can access models and datasets from the Hugging Face Hub\footnote{\url{https://huggingface.co/docs/hub/index}}.
Beyond corpora and models, \wb{} also incorporates code from multiple sources.

\section{Architecture}
\label{sec:architecture}

\begin{figure*}[th]
    \includegraphics[width=\linewidth]{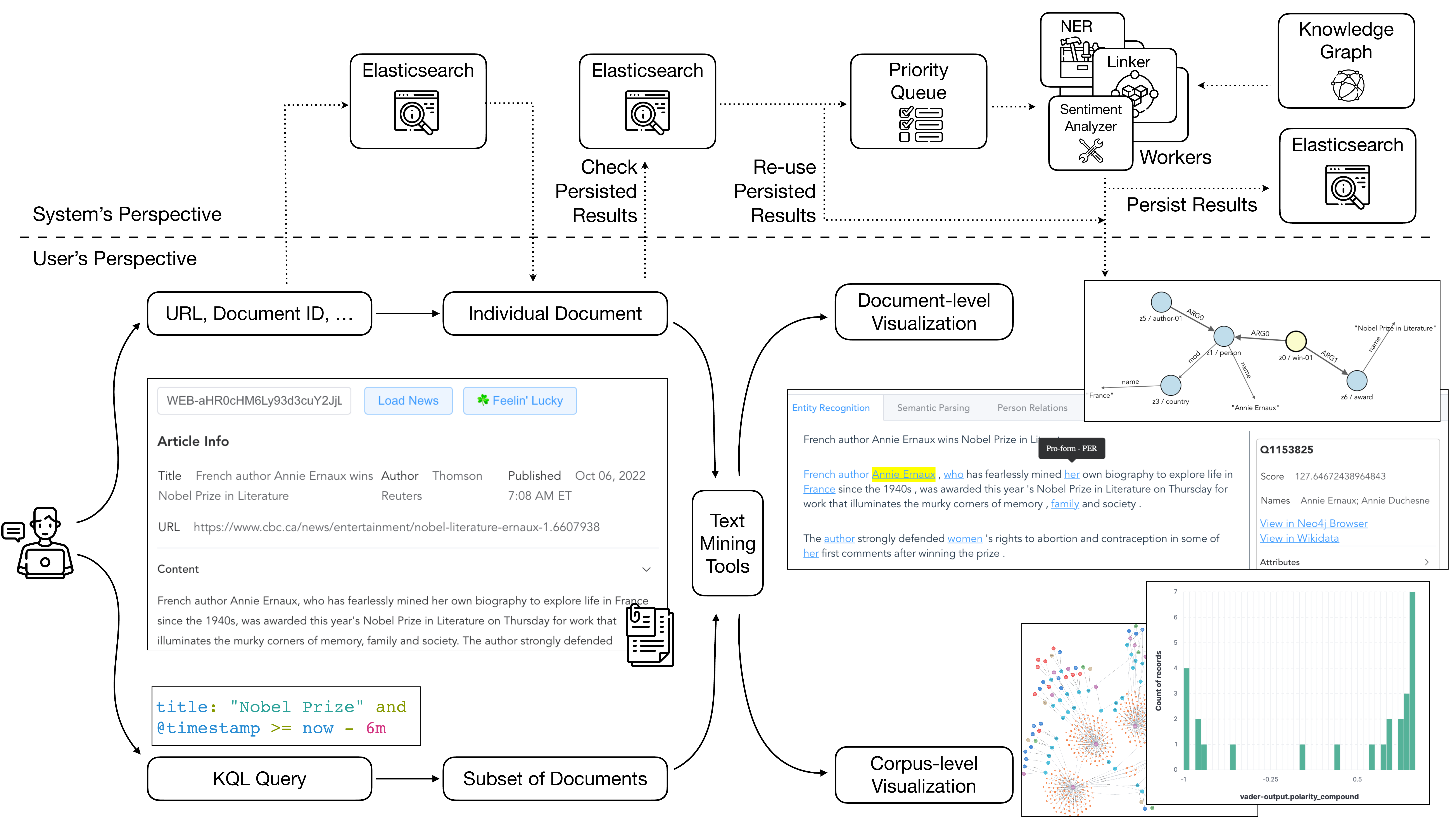}
    \caption{Workflow of \wb{} from the perspectives of the user and the system, as described in \S\ref{sec:workflow}.
    For document-level visualization, we showcase the user interface for named entity recognition, coreference resolution, entity linking, and semantic parsing.
    For corpus-level visualization, this figure includes plots of a social network constructed from a tweet of the official Nobel Prize account, and the distribution of sentiment polarity scores of a sample of the built-in news corpus.
    Icons created by Freepik - Flaticon.
    }
    \label{fig:workflow}
\end{figure*}

\wb{} is built on top of various open source software and incorporates the code and models from many research projects.
We design our architecture to leverage off-the-shelf functionalities provided by these software and projects, and to minimize the effort of integrating new code from a research project.

\subsection{Workflow}
\label{sec:workflow}
Figure~\ref{fig:workflow} provides a high-level overview of how users interact with \wb{} and how the system handles the requests.

\paragraph{User Perspective}
From the perspective of a user, one could choose to apply text mining tools on a single document specified by its URL or ID, or apply them in batches on a set of documents specified by a query.
Queries are written in the Kibana Query Language (KQL)\footnote{\url{https://www.elastic.co/guide/en/kibana/current/kuery-query.html}}, which is a simple and intuitive text-based query language.
The outputs of the tools on a single document can be visualized in the web interface, with each tool having a separate panel.
Using Kibana Lens\footnote{\url{https://www.elastic.co/kibana/kibana-lens}}, a user can visualize statistics calculated over the output of multiple documents, such as the distribution of sentiment polarity scores.
The connected Neo4j Browser\footnote{\url{https://neo4j.com/developer/neo4j-browser/}} provides an interactive web interface for exploring social networks constructed from a corpus.

\paragraph{System Perspective}
From the perspective of the system, both the corpus and the outputs of text mining tools are stored and indexed in Elasticsearch\footnote{\url{https://www.elastic.co/what-is/elasticsearch}}, a document indexing, search, and analytics engine.
By storing and indexing tool outputs, we re-use previous results and avoid re-computation to improve efficiency.
In addition to that, Elasticsearch provides convenient tools to filter documents based on the outputs of text mining tools and visualizing statistics, which are very useful for downstream analytics.

If running a tool is indeed necessary, the task is added to a priority queue.
Ad hoc and interactive requests, issued when a user is examining a single document and applying tools on it, are prioritized over batched requests that run in the background.
Each tool or model has workers processing the tasks in the queue.
This ensures that users performing interactive analysis experience little latency even when the number of workers is limited, which is usually the case in practice as deep learning models are often resource-intensive and it is infeasible to have multiple instances running in parallel.

\begin{figure*}[th]
    \includegraphics[width=\linewidth]{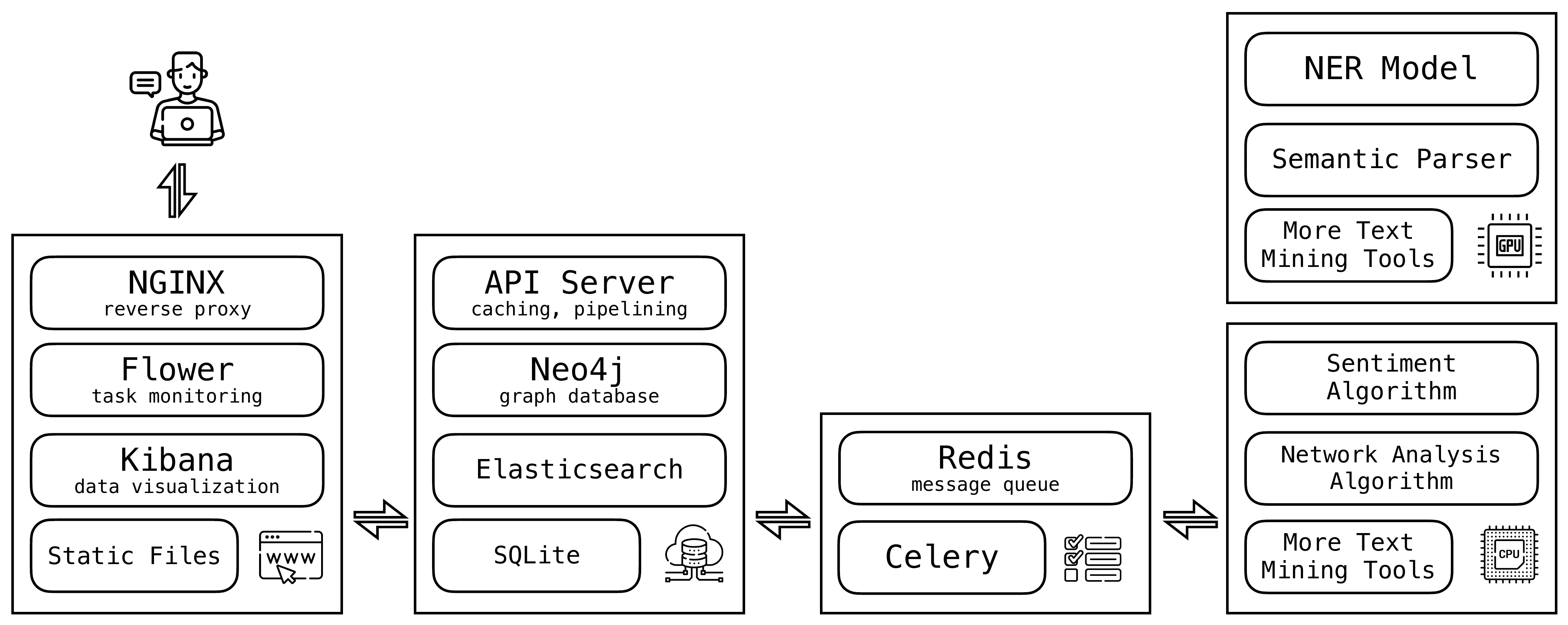}
    \caption{Microservice architecture of \wb{}.
    Each rectangle represents a physical machine, with its capability indicated by the icon at the bottom right corner.
    Each rounded rectangle represents a container, with the tool and function it provides indicated by the text inside.
    Container to physical machine allocation is for illustration purposes only and is adjusted to fit the need when the system is deployed in production.}
    \label{fig:containers}
\end{figure*}

\subsection{Pipelining and Scheduling}
\begin{figure}[ht]
    \includegraphics[width=\linewidth]{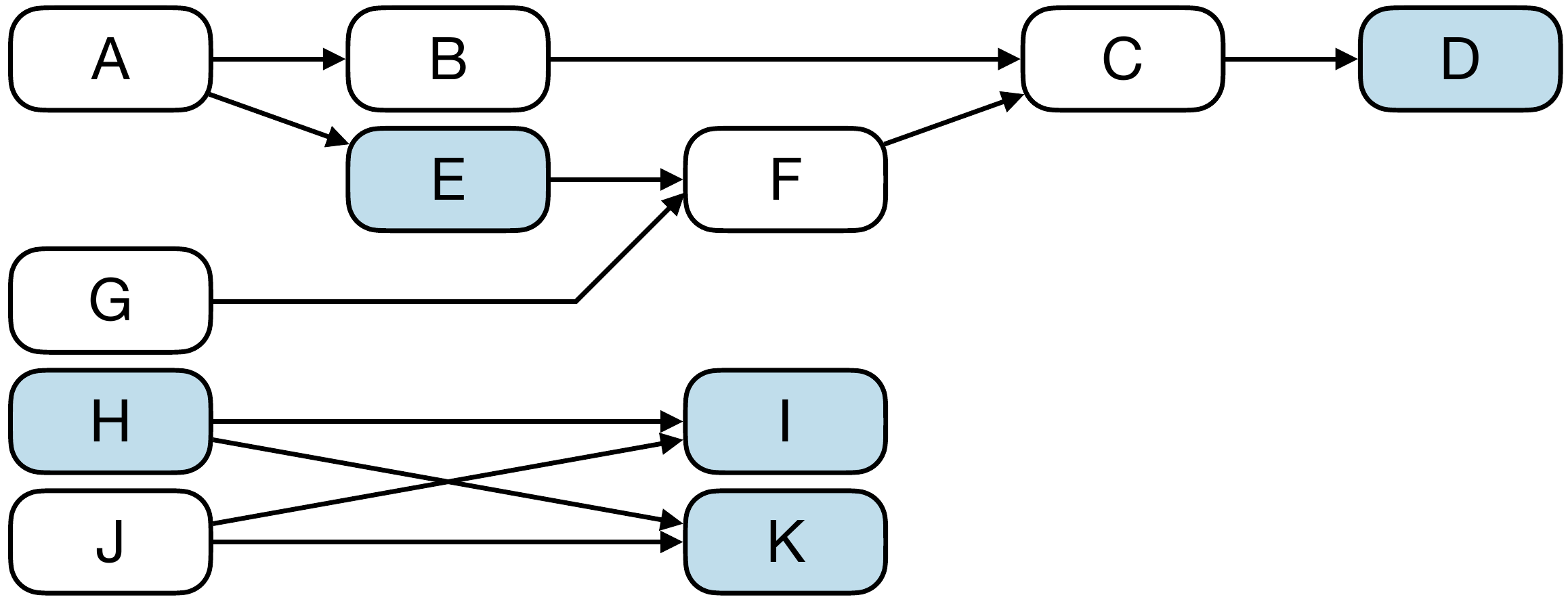}
    \caption{Example of a batched task with the directed acyclic graphs of dependencies. Shaded nodes represent tools that the user requests to run on the document, and unshaded nodes represent tools that are needed to provide the inputs to the shaded nodes.}
    \label{fig:dag}
\end{figure}

Text mining tools often rely on the outputs of other tools or NLP models and are built as pipelines.
For example, both entity linking and relation extraction require named entity recognition and coreference resolution.
To ensure efficiency, re-computing the outputs of tools that are already available should be avoided, and tools should be run in parallel if possible.
In addition to persisting and re-using outputs as discussed in Section \ref{sec:workflow}, we design a pipelining and scheduling system that automatically detects the dependencies between tools and schedules the tasks in a way that eliminates re-computation and encourages parallelism.

The dependencies of a tool can naturally be represented as a directed acyclic graph (DAG), where inbound edges represent the dependencies.
When multiple tools are requested to be run on a single document, we gather the direct and transitive dependencies of these tools in a single graph, as shown in Figure~\ref{fig:dag}.
The connected components of the graph are DAGs.
Within each DAG, the tools are run in topological order; and disjoint DAGs are executed in parallel. 

In the example illustrated in Figure~\ref{fig:dag}, the user requests to run tool \texttt{E, D, H, I}, and \texttt{K} on a document. The scheduler will automatically find all dependencies (\texttt{A} to \texttt{K}) and run two chains in parallel: \texttt{A-B-E-G-F-C-D} and \texttt{H-J-I-K}\footnote{There is more than one valid topological sorting for a DAG.}.

\subsection{Containerized Microservices}

A major obstacle to integrating third party code is the \textit{dependency hell} problem: it is an NP-complete problem to find a set of compatible versions of all software library dependencies \cite{burrows2005modelling,cox2016version}, and in reality a compatible set may not exist.
This is especially true for deep learning models \cite{han2020empirical,huang2022demystifying}, which often require specific versions of software libraries.
In the case of popular deep learning frameworks, TensorFlow \cite{tensorflow2016whitepaper} 2.0 introduces breaking API changes that are not back-compatible. Both TensorFlow and PyTorch \cite{paszke2019pytorch} are compiled with specific versions of CUDA \cite{nickolls2008cuda} and cuDNN \cite{chetlur2014cudnn}, and that makes different framework versions hard to coexist.
Manually fixing the code to make it compatible with a specific version of a library is often tedious and error-prone \cite{han2020empirical}.

For deployment, a practical problem is that it is often difficult or costly to find a single physical machine that satisfies the computing and networking requirements of all the components:
deep learning models require GPU for inference, database management systems consume large amounts of memory and disk space, and web servers need access to the Internet.
One solution to this problem is the ability to deploy components of \wb{} on multiple machines, which is achieved by our design.

We solve both problems discussed above at once by deploying both text mining tools and infrastructure components as \textit{containerized microservices}.
Each component is deployed as a Docker container \cite{merkel2014docker} that encapsulates the software and its dependencies. The containers communicate with each other via the RPC and message queue functions provided by Celery\footnote{\url{https://docs.celeryq.dev}}.
Figure~\ref{fig:containers} illustrates how the containerized microservices are deployed on separate machines with different capabilities.
Such a microservice architecture allows us to overcome the problem of heterogeneous technologies and simplifies horizontal scaling when needed \cite{newman2015microservices}.
\section{Components}
\label{sec:components}
\wb{} already includes a variety of tools and models for text mining.
Most of the components come from state-of-the-art research in the respective subfields. Others are baseline implementations to demonstrate \wb{}'s extensibility, showing that developers can straightforwardly incorporate new tools and build pipelines from existing ones.
One benefit of the flexible and modular design as described in \S\ref{sec:architecture} is that all built-in tools and models can easily be replaced or upgraded.
Existing tools and models in \wb{} include:

\paragraph{Named Entity Recognition}
The task, known as NER, is to identify mentions to entities such as people, organizations, and locations.
We incorporated the NER model from PURE \cite{zhong2021frustratingly}, which achieved good performance by simply fine-tuning BERT \cite{devlin2019bert}.

\paragraph{Coreference Resolution}
To determine which entity a pronoun refers to, we adopted the heuristic algorithm by \citet{cunningham2002gate} that is based on recency and type agreement.

\paragraph{Entity Linking}
Mentions to entities in the text are disambiguated and linked to Wikidata \cite{vrandecic2014wikidata} entities.
Candidate entities are generated by a fuzzy match on name.
In addition to name similarity, the ranking of candidates utilizes the cosine similarity between the sentence embeddings \cite{reimers2019sentence} of the context and the descriptions of the candidate entity from Wikipedia and Wikidata.

\paragraph{Relation Extraction}
The user can extract structured facts in the form of knowledge triples like (\textit{Annie Ernaux}, \texttt{Country}, \textit{France}) from a text. The underlying model \cite{mesquita2019knowledgenet} combines syntax and semantic features as well as BERT embeddings to predict the relation between entities.

\paragraph{Semantic Parsing}
Semantic parsing provides a structured representation of the meaning of a sentence, allowing users to obtain information like who did what to whom, when, and where without caring about the form.
\wb{} uses AMRBART \cite{bai2022graph}, a sequence-to-sequence model based on BART \cite{lewis2020bart} and pretrained on a large graph corpus, to parse sentences into AMR graphs \cite{banarescu2013amr}.

\paragraph{Summarization}
We build an application on top of semantic parsing to create natural language summaries of events related to people in the document, partly to demonstrate the simplicity of building pipelines in \wb{}.
For each sentence in the document, we prune its AMR graph to only contain the nodes and edges of pattern \textit{subject-predicate-object}, where the subject or object is a person.
The pruned AMR graphs are then converted to natural language using AMRBART.

\paragraph{Sentiment Analysis}
The sentiment of a document is predicted by VADER \cite{hutto2014vader}, a fast and accurate rule-based algorithm optimized for social media posts.
A sentiment polarity score is produced and can be used to classify the sentiment as positive, neutral, or negative.

\paragraph{Social Network Analysis}
For corpora consisting of social media posts, \wb{} is equipped with a tool that builds graphs of social network interactions from posts.
Powered by the graph database Neo4j\footnote{\url{https://neo4j.com/}}, the tool can be used to visualize the network and perform analyses, such as running centrality algorithms like PageRank \cite{page1999pagerank} to identify influential users.
\section{Use Cases}
\label{sec:use-cases}
Text mining has been proven useful in a variety of domains, such as corporate finance, patent research, life sciences, and many others \cite{feldman2006applications}.
\wb{}, as a full-fledged text mining platform, has been or has the potential to be applied in many of these domains.
Some of the use cases are described in this section.

\paragraph{Digital Humanities}
Accessible and reliable NLP tools are useful in digital humanities projects such as Linked Infrastructure for Networked Cultural Scholarship (LINCS)\footnote{\url{https://lincsproject.ca}}. A major component of the LINCS project is to generate linked data from natural language cultural heritage texts. The key steps are NER, coreference resolution, entity linking, and relation extraction, all of which are made available in \wb{}. Through the simple user interface, digital humanists and their students can choose the available models that suit their data, and easily connect these steps into a custom workflow.

The Centre for Artificial Intelligence, Data, and Conflict (CAIDAC)\footnote{\url{https://www.tracesofconflict.com/}} houses human right scholars interested in how armed groups and extremists use social media to promote their agendas. \wb{} offers such researchers convenient tools for collecting  tweets\footnote{Support for Telegram and other platforms is in progress.} for analysis. Posts can be subjected to the NLP tools of interest, and the social network underlying the corpus can be visualized, explored, and analyzed. All of these tasks are done through an accessible interface, requiring no programming from the users, enabling them to perform more and larger studies in a fraction of the time otherwise required.

\paragraph{Business Analytics}
Business analysts ask questions like ``\textit{are recent news reports about Apple Inc. positive or negative?}''.
These type of questions can easily be answered by \wb{}.
After performing NER and entity linking on the news articles, the analyst can conduct a \textbf{semantic search} to find the articles that are related to \textit{Apple Inc.} rather than apple the fruit.
Then, the analyst can use the sentiment analysis tool and visualize the distribution of sentiment polarity scores with Kibana Lens, as shown in the bottom right screenshot in Figure~\ref{fig:workflow}.

\paragraph{NLP Research}
All NLP models in \wb{} can be accessed via RESTful API and RPC, or used directly as containers.
For researchers who wish to perform inferences with the models on their own data, they could use the interfaces provided by \wb{}, without needing to set up the environment to run the models.
\section{Roadmap}
\label{sec:roadmap}
\wb{} is still in its early stages of development, and we are actively working on improving the system.
Besides usability, stability, and security updates, we plan to work on the following major features in the near future:

\paragraph{Human-in-the-loop NLP}
Adding annotation support to the web interface will allow users to provide feedback to the outputs of models.
This will help researchers to collect domain-specific labelled data and improve the performance of the models in a human-in-the-loop fashion \cite{wang2021putting}.

\paragraph{Improved Corpus Management}
Managing document collections is a crucial aspect of text mining \cite{hearst1999untangling,cunningham2002gate}.
Currently, corpora are manually imported into Elasticsearch or created by crawling social media.
We hope to improve the way users access document collections.
This can be done by connecting \wb{} to Datasets \cite{lhoest2021datasets} and DataLab \cite{xiao2022datalab} where popular text datasets are already available.
In addition to crawling from social media, we also plan to support creating a new corpus by doing web search on a search engine.

\paragraph{More Text Mining Tools}
The extensible design of \wb{} allows us to keep existing tools and models up to date by replacing them when better models are released, and integrate emerging text mining tools to the system.
For example, we hope to add claim extraction models to facilitate fact checking tasks \cite{hassan2017claimbuster}.

\paragraph{Multi-modal Analysis}
Social media posts often refer to or contain information in other modalities (images, video, audio) of interest. At the same time, there is growing interest in grounding NLP models and analysis on knowledge extracted from videos and other sources. While adding support for processing different media in \wb{} is as easy as adding more NLP tools, we are interested in integrating these models so that co-training or grounding can be automated to the extent possible.
\section{Conclusion}
We introduced \wb{}, a platform that caters to all three major aspects of text mining systems: corpus management, text mining tools, and user interface. We explained what design features make \wb{} efficient and extensible, and how it can be used in a variety of applications.

\section*{Limitations}
We have already identified several important features that are not yet implemented in \wb{}, as discussed in \S\ref{sec:roadmap}: the platform needs an annotation feature for human-in-the-loop AI; it should have access to commonly used public corpora; and it should include text mining tools such as one for claim extraction.
There are some intrinsic limitations that even the state-of-the-art models in \wb{} do not solve.
For example, long tail entities may not be covered by the knowledge graph, and current entity linking models do not have a notion for out-of-knowledge-graph entities \cite{shen2021entity}.
This will result in long tail entities always being incorrectly linked.
Beyond social network analysis, our current design does not have the user interface or models for other corpus-level analyses, such as topic modeling \cite{blei2003lda}.
And finally, all NLP models and algorithms in \wb{} are targeted at English text. Although we have been able to deal with corpora in other languages by translating them to English using Marian MT \cite{junczys2018marian}, it is not yet clear whether performance can be improved by directly using models trained on other languages.

\section*{Ethics Statement}
By encapsulating the models, \wb{} lowers the entry barrier for non-experts to use state-of-the-art AI models.
The microservice architecture, which allows models to be deployed on multiple servers with different capabilities rather than a single omnipotent server, also makes this text mining platform more accessible.
Containerizing third-party models also helps with reproducibility and transparency.
There have been attempts of using \wb{} to analyze datasets to help understand propaganda, misinformation, and disinformation related to war and terrorism.
However, users must be warned that, as \wb{} uses third-party data and models without modification, outputs obtained from \wb{} are inevitably affected by the bias inherent in the datasets and models.

\section*{Acknowledgements}
The work is supported in part by the Natural Sciences and Engineering Research Council of Canada (NSERC), the Canadian Foundation for Innovation (CFI), AI4Society\footnote{\url{https://ai4society.ca/}} and a gift from Scotiabank.
Certain computing resources are provided by the Digital Research Alliance of Canada.

\bibliography{custom}
\bibliographystyle{acl_natbib}

\end{document}